\pdfoutput=1

\documentclass[11pt]{article}

\usepackage[final]{EMNLP2023}

\usepackage{times}
\usepackage{latexsym}

\usepackage[T1]{fontenc}

\usepackage[utf8]{inputenc}

\usepackage{microtype}

\usepackage{inconsolata}
\usepackage{amsmath}
\usepackage{wasysym}
\usepackage{multirow}
\usepackage{graphicx}
\usepackage{hyperref}
\usepackage{longtable}

\title{Making FETCH! Happen: Finding Emergent Dog Whistles Through Common Habitats}

\author{
 \textbf{Kuleen Sasse\textsuperscript{1}},
 \textbf{Carlos Aguirre\textsuperscript{1}},
 \textbf{Isabel Cachola\textsuperscript{1}},\\
 \textbf{Sharon Levy\textsuperscript{2}},
 \textbf{Mark Dredze\textsuperscript{1}}
\\
 \textsuperscript{1}Johns Hopkins University,
 \textsuperscript{2}Rutgers University
\\
 \small{
   \textbf{Correspondence:} \href{mailto:ksasse1@jh.edu}{ksasse1@jh.edu}
 }
}

\begin{document}
\maketitle
\begin{abstract}
\textcolor{red}{WARNING: This paper contains content that maybe upsetting or offensive to some readers.}\\
Dog whistles are coded expressions with dual meanings: one intended for the general public (outgroup) and another that conveys a specific message to an intended audience (ingroup). Often, these expressions are used to convey controversial political opinions while maintaining plausible deniability and slip by content moderation filters. Identification of dog whistles relies on curated lexicons, which have trouble keeping up to date. We introduce \textbf{FETCH!}, a task for finding novel dog whistles in massive social media corpora. We find that state-of-the-art systems fail to achieve meaningful results across three distinct social media case studies. We present \textbf{EarShot}, a strong baseline system that combines the strengths of vector databases and Large Language Models (LLMs) to efficiently and effectively identify new dog whistles.
\footnote{Code for analysis and evaluation can be found at \texttt{https://github.com/KuleenS/FETCH-Dog-Whistle}} 

\end{abstract}

\section{Introduction}

\begin{quote}
    {
    ``\textit{white house petition: end \textbf{dual citizens} from serving in us and state governments}'' \\\phantom{abc}--- \textbf{Anonymous Gab User}
    }
\end{quote}

When a person writes \textbf{dual citizens} it may seem like an innocuous phrase, but many understand this as an antisemitic reference. The use of the phrase builds on centuries old questions of the loyalty of Jewish people towards governments, which recently has become a claim that all American Jews are double citizens of Israel and the United States.

This phrase is a dog whistle, which is defined as an ``expression that sends one message to an outgroup while at the same time sending a second (often taboo, controversial, or inflammatory) message to an ingroup \cite{henderson2018dogwhistles}.'' Dog whistles are a common mainstay in today's turbulent political climate. For example, they prominently featured in the 2016 United States Presidential Election both on social media and in person \cite{tilley2020law, drakulich2020race}.  

\begin{figure}
    \centering
    \includegraphics[width=\linewidth]{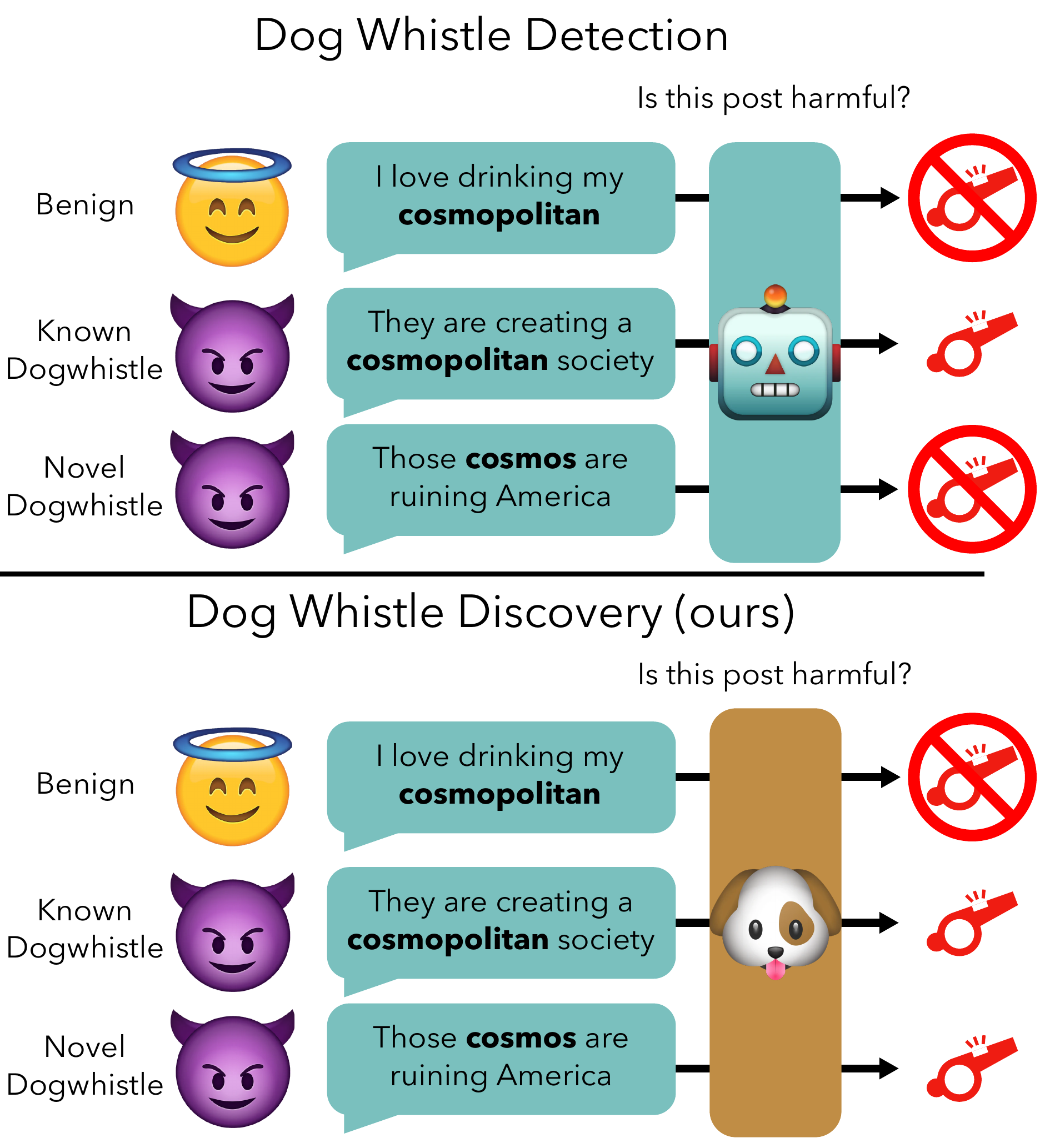}
    \caption{Comparison between dog whistle detection and our method of dog whistle discovery}
    \label{fig1:dogwhistle-explanation}
\end{figure}

As dog whistles are a common way to transmit hate speech to a wide audience both digitally and offline, researchers have started to develop methods to detect them. However, dog whistles are notoriously difficult to find as they often slip by content moderation, toxicity, and hate speech filters at a high rate because of their benign second meaning \cite{mendelsohn-etal-2023-dogwhistles}. Dual citizenship isn't an issue, but ``dual citizen'' in the above context has a specific meaning. More examples with explanations of the dog whistles' meaning and their difficulty are provided in Appendix \ref{appx:dog-whistles-examples}. 

Currently, detecting dog whistles relies on manually curated lexicons, which can be effective but are labor-intensive to create and maintain. They cannot adapt to the dynamic evolution of language. For example (Figure \ref{fig1:dogwhistle-explanation}), current methods can distinguish between the two meanings of the word ``cosmopolitan'': its use as the popular mixed drink and its use as a dog whistle to denigrate a multicultural and intellectual society. However, they miss a hypothetical linguistic shift of the term ``cosmopolitan'' to the shortening of ``cosmos.'' 

We consider the task of dog whistle discovery, a novel task where systems identify emerging or previously unknown dog whistles. To support this effort,  we create \textbf{FETCH!} or \textbf{F}inding \textbf{E}mergent Dog Whistles \textbf{T}hrough \textbf{C}ommon \textbf{H}abitats, a benchmark designed to evaluate models' ability to detect emerging or novel dog whistles across diverse settings. Using the \textbf{FETCH!} benchmark, we evaluate three state of the art methods across three different social media case studies. Our evaluation reveals that these methods perform poorly, achieving an F-score of less than 5\% across all case studies.

We introduce our own method as a strong baseline: \textbf{EarShot}. This method finds new dog whistles by leveraging the power of sentence embeddings and Large Language Models (LLMs). EarShot demonstrates significant improvements, with F-score increases ranging from $2$ to $20$ points across the social media case studies. We make the following key contributions: Development of \textbf{FETCH!} benchmark for dog whistle discovery; Rigorous evaluation of multiple state-of-the-art methods on \textbf{FETCH!}; Creation of \textbf{EarShot}, a strong baseline architecture achieving SoTA performance on \textbf{FETCH!};

\section{Related Work}

\subsection{Implicit Hate Speech Detection}
Detecting novel dog whistles can be thought of as a form of hate speech detection, specifically implicit hate speech detection. Implicit hate speech can be defined as hate speech that uses coded or indirect language \cite{waseem-etal-2017-understanding}. Many studies have focused on this task through creating new datasets \cite{hartvigsen2022toxigenlargescalemachinegenerateddataset, elsherief-etal-2021-latent, ocampo-etal-2023-depth, breitfeller-etal-2019-finding}, novel developing detection methods \cite{zhang2024dontextremesrevealingexcessive, gao-etal-2017-recognizing}, and methods to explain the hate speech \cite{qian-etal-2019-learning}. However, these methods focus on the overall text or post rather than the specific words and phrases. 
  
\subsection{Dog Whistles}
Other fields like linguistics, political science, and other social sciences have researched the use of dog whistles and their appearance through various different forms of analysis. Linguistic methods have focused on the semantics and pragmatics of their use \cite{henderson2018dogwhistles, 48062b90-c40f-3ea1-8ec1-2b3cf1974995, dc901c7f-3053-377d-8b63-ed6f6a8167af, saul2018dogwhistles, bhat2020covert, quaranto2022dog} and further extended their work using agent-based simulations of the emergence of new dog whistles \cite{henderson-mccready-2020-towards}. Historians have focused on showing the evolution and propagation of racial and dog whistle speech for political purposes and their lasting effects on society \cite{mendelberg2001race, haney2014dog}. Sociologists and psychologists have focused on surveying attitudes of people before and after reading messages with implicit racial appeals like dog whistles \cite{wetts2019called, albertson2015dog}. Other sociologic work has studied the development and changes of dog whistles over time on the web \cite{aakerlund2022dog}.  Political science has investigated the use of dog whistles and how they change public opinions and policies surrounding controversial topics like sanctuary cities and other criminal justice policies \cite{lasch2016sanctuary, hurwitz2005playing, goodin2005dog}.

\subsection{Euphemism Detection} 
Euphemism detection focuses on finding novel euphemisms, terms or phrases used to substitute more offensive terms in order to downplay its unpleasantness, a semantic phenomenia similar to dog whistles. Typically, the task is structured as given a set of seed euphemisms, the goal of the task is to find other euphemisms in a provided corpora, a process similar to lexicon induction \cite{lee-etal-2022-report}. 

Early works in euphemism detection often focused on using Word2Vec \cite{mikolov2013efficientestimationwordrepresentations} to find euphemisms. In one work, the authors trained a Word2Vec model on the corpus of interest, and then, they used the similarities between the word vectors to obtain similar words to their seed words \cite{magu-luo-2018-determining}. Other works further improved on this method combining both Word2Vec and dependency based word embeddings \cite{levy-goldberg-2014-dependency, taylor2017surfacingcontextualhatespeech}, using two different Word2Vec models on two different datasets and finding the words with the greatest difference in embeddings \cite{9231109,yuan2018reading, 7745450}, or using the difference between the two Word2Vec learned matrices \cite{aoki-etal-2017-distinguishing}. Other works used methods combining Word2Vec methods with additional user and search behaviors of users \cite{7958608}. Later works further refined these methods by using both more powerful Masked Language Models like BERT \cite{devlin2019bertpretrainingdeepbidirectional, zhu2021selfsupervisedeuphemismdetectionidentification} and SpanBERT \cite{joshi2020spanbertimprovingpretrainingrepresenting, zhu-bhat-2021-euphemistic-phrase} both in a zeroshot setting \cite{maimaitituoheti-etal-2022-prompt} and a supervised setting and Large Language Models like GPT-3 \cite{brown2020languagemodelsfewshotlearners} in a zeroshot setting \cite{keh-2022-exploring}.  

\subsection{Dog Whistle Detection}
Relatively few studies have addressed the detection of dog whistles. Among these, three notable works focus specifically on Swedish dog whistles, employing both embedding-based methods \cite{boholm-etal-2024-political, hertzberg-etal-2022-distributional} and analyzing semantic shifts \cite{boholm-sayeed-2023-political-dogwhistles} In English, only two studies have focused on detecting dog whistles. In the first paper, they curated a set of English dog whistles to assess both large language models’ (LLMs) ability to surface these terms and the impact on hate speech detectors. However, they only do a small analysis with a handful of positive examples to see if LLMs can detect novel dog whistles \cite{mendelsohn-etal-2023-dogwhistles}. In the second paper, they introduced the largest dataset of posts containing dog whistles, filtered from sources such as Reddit and the Congressional Record by LLMs. In addition to the creation of the dataset, they conducted a small preliminary test of LLMs’ capabilities in detecting novel dog whistles \cite{kruk-etal-2024-silent}.

\section{Task Description}
We propose a novel task: \textbf{Finding Emergent Dog Whistles in Common Habitats} or \textbf{FETCH!}. For the task, a system is provided with a corpus and a set of initial seed dog whistles. The corpus might be social media posts, forums, website pages, etc. The initial set of seed dog whistles is a list of known dog whistles that occur within the corpus. The goal of the task is to use the initial seed words and the corpus to discover other dog whistles.

\subsection{Case Studies}
We describe the three example case studies that represent real-world or synthetic scenarios for our task.  

\subsubsection{Synthetic (Reddit)}
Our first scenario is an idealized setting where every post contains an associated dog whistle. We use the Silent Signals dataset \cite{kruk-etal-2024-silent}, which includes 16,000 posts scraped from Reddit from 45 controversial subreddits. Each post in the dataset contains a dog whistle identified by GPT-4 \cite{openai2024gpt4technicalreport}.

\subsubsection{Balanced (Gab)}
The balanced scenario represents a context with a higher-than-average prevalence of dog whistles with large amount of posts. For this scenario, we chose to use data from Gab, an alt-right alternative to Twitter where dog whistles and other forms of hate speech are more frequent. The data was collected over a short period using a web scraper that sequentially downloaded post IDs yielding approximately 300,000 posts.

\subsubsection{Realistic (Twitter)}
The realistic scenario reflects a typical online environment where dog whistles are sparse, and most posts are benign. For this scenario, we selected X (formerly Twitter) as the platform, as it contains some hate speech but removes much of it through content moderation. Data were collected using the $1$\% public Twitter API stream between January 2017 and December 2022. Given the dataset's massive size (approximately $1$ billion tweets), we subsampled it to make it manageable. We randomly sampled 6 million English language tweets without dog whistles and $1$ million tweets containing dog whistles identified using regular expression from a set of ground truth dog whistles. This process resulted in a dataset of 7 million tweets.

\subsection{Seed Dog Whistles}
We use the dog whistle dataset released by \citet{mendelsohn-etal-2023-dogwhistles} as our ground truth.
The dataset is a hand-curated taxonomy of around 340 root dog whistles from community wikis and academic sources. Each root dog whistle comes other surface forms of the root dog whistle. We use these dog whistles for the Realistic and Balanced scenarios as the Synthetic scenario has its own set of dog whistles.
Table \ref{tab:highest_possible} lists a summary of the number of dog whistles found in each scenario. The full list appears in Appendix \ref{appx:scenario-dogwhistles}.

\begin{table}[!ht]
\resizebox{\columnwidth}{!}{%
\begin{tabular}{l|rrr}
Scenario                  & \multicolumn{1}{l}{Balanced} & \multicolumn{1}{l}{Synthetic} & \multicolumn{1}{l}{Realistic} \\ \hline
Total Dog Whistles Found & 77                           & 298                           & 177                           \\ \hline
Reference Dog Whistles Total       & 340                          & 298                           & 340                           \\ \hline
\end{tabular}%
}\caption{Number of dog whistles root forms found in each scenario}
\label{tab:highest_possible}
\end{table}

For each corpora, we divide the dog whistles that are able to be found into train/test splits (20\% / 80\%), creating a set of seed dog whistles to use to find the rest of the dog whistles. We split the dog whistles using stratified sampling based on the length of phrase in $n$-grams. We do a stratified split so as not to unfairly punish methods that can identify larger $n$-grams. After splitting, we add the surface forms as well to each split in order to cover all variants of the dog whistles. We split on $n$-grams over different metadata provided in the dog whistle list like the group being targeted or the user of the dog whistle as the Synthetic scenario does not provide this information and it is more realistic scenario to have a set of randomly chosen dog whistles rather a well distributed set of dog whistles according to the metadata. 

\subsection{Metrics for Evaluation}
We use precision, recall and F-score to measure the performance of different methods for our task. We opt to use precision, recall, and F-score without a threshold as it is a more general metric as some methods do not rank their predictions. In addition, we do not create metrics that normalize for the maximum length of the words output as in testing, models with smaller length outputs have inflated scores. 

Due to the multiple surface forms of dog whistles provided in the dataset, all metrics count a positive match for a single root dog whistle as returning an exact match of one of the surface forms or the root itself. 

Precision is computed as normal, but we modify recall to normalize by only the positive terms that exist in the corpus (Data Potential Recall, DPR). We report $F_{0.5}$ scores using DPR and Precision instead of $F_1$ to preference methods with a high precision to reduce the load on a human reviewer.

\section{Experimental Setup}
\subsection{Models}
We test four different methods on \textbf{FETCH!}: Word2Vec/Phrase2Vec \cite{mikolov2013efficientestimationwordrepresentations}, MLM \cite{zhu2021selfsupervisedeuphemismdetectionidentification}, EPD \cite{zhu-bhat-2021-euphemistic-phrase}, and our method: EarShot. 
\subsubsection{Word2Vec/Phrase2Vec}
Word2Vec and Phrase2Vec are widely used models for euphemism detection due to their lightweight nature and fast computation. These models capture semantic relationships between words or phrases, enabling the detection of euphemistic language. In our implementation of these models for dog whistle discovery, we adopt a methodology similar to \cite{magu-luo-2018-determining}. 

To create the training data for the Word2Vec/Phrase2Vec \cite{mikolov2013efficientestimationwordrepresentations}, we preprocessed the corpus by removing all stop words and lemmatizing the rest using \texttt{littlebird} \cite{DeLucia2020}. After this preprocessing step, we replaced URLs, hashtags, retweets, and mentions with special symbols like @USER and HTTPURL. We also replaced emojis with their official English aliases using the \texttt{emoji} package \cite{emoji}. Finally, we tokenized the result with the NLTK Tweet Tokenizer \cite{bird-loper-2004-nltk}. 

To construct the Phrase2Vec models, we utilized the Gensim Phraser module \cite{rehurek2011gensim} to merge common $n$-gram phrases in the original corpus, generating $2$-gram and $3$-gram versions of the models. We only train up to $3$-gram models as a majority of dog whistles in our dataset are $3$-grams and smaller. We then trained models using Gensim with a maximum vocabulary size of 500,000 and a context window of 5 with a vector dimension of 100 for 10 epochs.

To generate the predictions, we provided the Word2Vec/Phrase2Vec model the list of seed dog whistles and retrieved the top 10 most similar words to each seed word. Then, we queried for the top 10 most similar words to those words. We continued this process 10 times creating 10 levels of predictions. 

\subsubsection{MLM}\label{section:MLM}
Euphemism detection relies heavily on context, but static embeddings from Word2Vec/Phrase2Vec fail to capture this. To address this, researchers leveraged contextual embeddings from Masked Language Models like BERT \cite{zhu2021selfsupervisedeuphemismdetectionidentification}. The MLM system first samples 2000 sentences from the input dataset containing the seed dog whistles. We sampled 2000 sentences as it is the default parameter provided. For each sampled sentence, it masks out the seed dog whistle in that sentence and uses the Masked Language Model to fill in the mask by creating a distribution over possible replacements and then returning the top-$k$ most likely. We test this method using two MLM encoder models: Bernice \cite{delucia-etal-2022-bernice} and BERTweet \cite{nguyen-etal-2020-bertweet}. Both these models have been adapted to social media data, providing a reasonable benchmark as pretraining custom models from scratch computationally expensive. We test the method at thresholds for $k$ of $\{50,100,200,400,800,1600,3200,6400,12800,\\25600\}$.

\begin{figure*}[!ht]
    \centering
    \includegraphics[width=\linewidth]{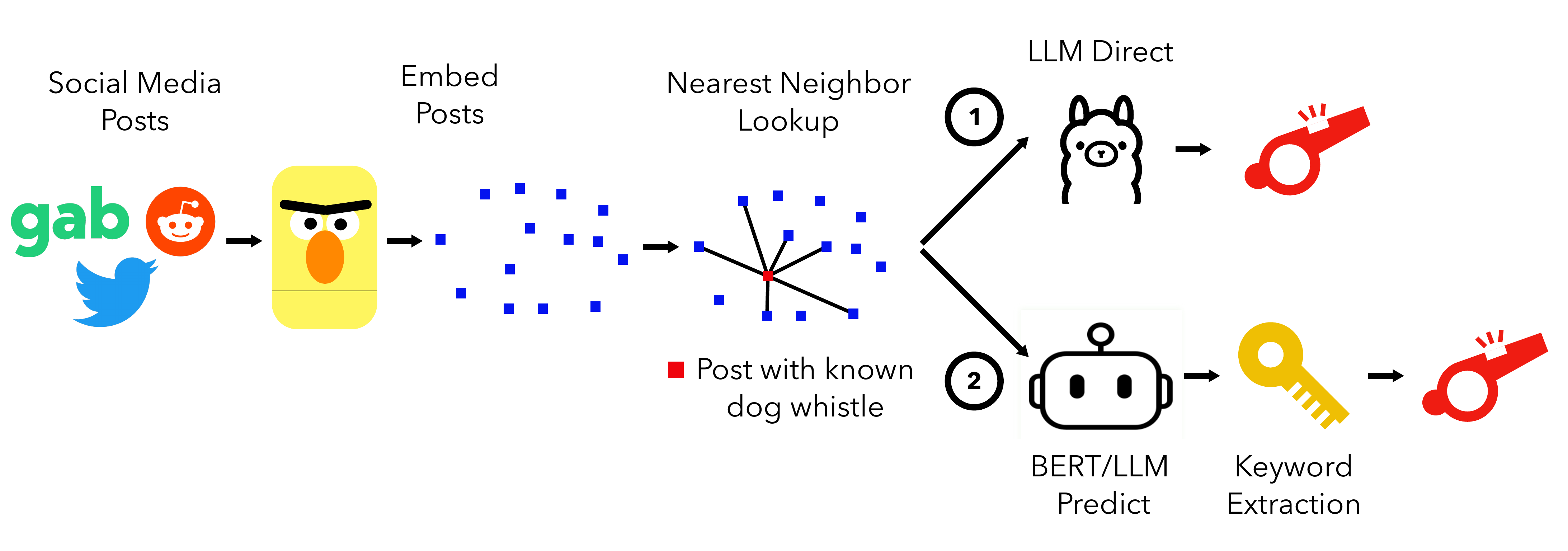}
    \caption{Flowchart for the EarShot System. Two paths can be taken after doing the nearest neighbor lookup. Path (1) is the more computationally expensive system that asks directly prompts a LLM for an answer. Path (2) is the cheaper method that leverages keyword extraction models and binary prediction. }
    \label{fig2:flowchart}
\end{figure*}

\subsubsection{EPD}\label{section:EPD}
The Euphemistic Phrase Detector (EPD) \cite{zhu-bhat-2021-euphemistic-phrase} builds upon the MLM model by predicting phrases instead of single tokens. The EPD model uses Autophrase \cite{shang2017automatedphraseminingmassive}, a statistical phrase extractor to extract important phrases from the corpus. Next, it trains a unigram Word2Vec model \cite{mikolov2013efficientestimationwordrepresentations} on the same corpus to filter down the phrases by selecting the top 1000 most similar phrases from AutoPhrase to the seed dog whistles. Finally, it samples 2000 sentences from the input dataset that contain the seed dog whistles. For each sampled sentence, it replaces the seed dog whistle with each predicted phrase from the previous stage and uses the Masked Language Model to calculate the probability of the sentence with the replaced phrase. They then sort the sentences via their probability and select the top-$k$ sentences and return the corresponding inserted phrase as predictions. We test this method using SpanBERT \cite{joshi2020spanbertimprovingpretrainingrepresenting} as our Masked Language Model as it is the default model used. We test the method at increasing thresholds for $k$ of $\{50,100,200,400,800,1600,3200,6400,12800,\\25600\}$.

\subsubsection{EarShot}
We propose EarShot, a strong baseline architecture for the task, consisting of three stages. 

In the first stage, all posts in the corpora are turned into vectors using a sentence transformer \cite{reimers-gurevych-2019-sentence} and put into a vector database. We opt to use the \texttt{all-MiniLM-L6-v2} as our sentence transformer as it is one of the fastest models with high quality embeddings. We use ChromaDB \cite{chroma} as our vector database. 

In the second stage, we obtain the vectors for the posts containing the seed dog whistles and find the closest vector to each post vector that are not the post themselves. By identifying closet neighbors, we aim to capture posts that are likely related in meaning, sentiment, or intent, even if they do not share exact word matches. After this step, we try two different methods: DIRECT and PREDICT, labeled (1) and (2) in Figure \ref{fig2:flowchart} respectfully. 

For the DIRECT pipeline, we pass all the posts to a LLM to extract the dog whistles from the posts provided from the vector lookup. The model is prompted to list all dog whistles in a structured JSON format with a prompt from \cite{kruk-etal-2024-silent}. The prompt is shown in Table \ref{tab:prompt_table} in Appendix \ref{appx:prompts}. We evaluate three LLMs: \texttt{meta-llama/Meta-Llama-3.1-8B-Instruct} (LLaMa 8B) \cite{dubey2024llama3herdmodels}, \texttt{mistralai/Mistral-7B-Instruct-v0.3} (Mistral 7B) \cite{jiang2023mistral7b}, and \texttt{meta-llama/Llama-2-13b-chat-hf} (LLaMa 13B) \cite{touvron2023llama2openfoundation}. We choose to test LLaMa and Mistral as they some of the best performing open source LLMs. We choose to use smaller open source models over larger closed source models as they are easier to reproduce and provide a strong baseline. 

For the PREDICT pipeline, we have two stages. In the first stage, we have a filtering step. We filter the posts using either a BERT model or a LLM. For the BERT based filtering, we use an off the shelf hate speech or toxicity detection model to classify the posts. We remove all the posts that are labeled not toxic or hate speech according to the model as dog whistles are commonly found in hate speech. We call this method filtering BERT-PREDICT. We test three popular hate speech and toxicity classifiers: \texttt{tomh/toxigen\_hatebert} (ToxiGen BERT) \cite{hartvigsen2022toxigenlargescalemachinegenerateddataset}, \texttt{facebook/roberta\-hate-speech-dynabench-r4-target} (RoBERTa R4) \cite{vidgen-etal-2021-learning}, and \texttt{Hate-speech\-CNERG/bert-base-uncased-hatexplain}  (HateXplain BERT)\cite{mathew2022hatexplainbenchmarkdatasetexplainable}. 

For the LLM based filtering, we prompt the model to reply yes or no if there is a dog whistle in the post. We structure our prompt similarly to Silent Signals \cite{kruk-etal-2024-silent}, shown in Table \ref{tab:prompt_table} in Appendix \ref{appx:prompts}. We keep all the posts that the model says yes to using a substring search. We call this filtering method LLM-PREDICT. We test the same three models used in the DIRECT pipeline.

After the filtering step, we pass the posts to a keyword extractors and return the top-$k$ predictions of them.  We use five different popular keyword extraction algorithms: KeyBERT \cite{grootendorst2020keybert}, RAKE \cite{rose2010automatic}, YAKE, \cite{campos2018text}, TextRank \cite{mihalcea-tarau-2004-textrank}, and TF-IDF \cite{sparck_jones_1972}. If the keyword extractor requires a specific range of $n$-grams to return, we test ranges of $1$-gram, $1$–$2$-grams, and $1$–$3$-grams. We report the performance of the method at the same thresholds in \hyperref[section:MLM]{MLM} and \hyperref[section:EPD]{EPD}. 

\section{Results}

\subsection{Word2Vec/Phrase2Vec}
\begin{table}[!ht]
\resizebox{\columnwidth}{!}{%
\begin{tabular}{l|lrrrr}
Scenario                   & Model   & \multicolumn{1}{l}{Level} & \multicolumn{1}{l}{Prec} & \multicolumn{1}{l}{DPR} & \multicolumn{1}{l}{$F_{0.5}$} \\ \hline
\multirow{3}{*}{Balanced}  & Unigram & 1                         & \textbf{1.64}            & 4.12                    & \textbf{1.86}            \\
                           & Bigram  & 1                         & 0.83                     & \textbf{6.17}           & 1.00                     \\
                           & Trigram & 1                         & 0.83                     & \textbf{6.17}           & 1.00                     \\ \hline
\multirow{3}{*}{Synthetic} & Unigram & 1                         & \textbf{5.50}            & 8.40                    & \textbf{5.91}            \\
                           & Bigram  & 2                         & 2.82                     & \textbf{21.85}          & 3.42                     \\
                           & Trigram & 1                         & 3.24                     & 7.98                    & 3.67                     \\ \hline
\multirow{3}{*}{Realistic} & Unigram & 1                         & \textbf{2.97}            & 1.47                    & \textbf{2.47}            \\
                           & Bigram  & 1                         & 0.89                     & 1.47                    & 0.97                     \\
                           & Trigram & 1                         & 1.03                     & 1.47                    & 1.10                     \\ \hline
\end{tabular}%
}\caption{Results for best Word2Vec/Phrase2Vec models by $F_{0.5}$. Prec is Precision. DPR is Data Potential Recall. Best scores for across each dataset are \textbf{bolded}}\label{tab:word2vec}
\end{table}

We summarize the best models for Word2Vec and Phrase2Vec in Table \ref{tab:word2vec}. Unigram models do the best across all three datasets both in terms of Precision and $F_{0.5}$ but it comes at the cost of lower recall as they cannot capture more complex phrases. 

On the Synthetic scenario, none of the models even hit $7$-$10$ F-score, which means they are predicting most words wrong even though all sentences have a dog whistle. On Realistic and Balanced, performance significantly drops as models can only achieve $1$-$2$ F-score showing how sparser environments hurt the Word2Vec model even more. 

Most models perform best at the first or second prediction level. This conservative strategy limits the model's usefulness, as it predicts only a small number of new dog whistles. Threshold choice figures are in Appendix \ref{sec:word2vec-app}. Overall, unigram Word2Vec models excel in this task but struggle to capture more complex phrases due to model limitations and their conservative prediction approach.

\subsection{MLM and EPD}
\begin{table}[!ht]
\resizebox{\columnwidth}{!}{%
\begin{tabular}{l|lrrrr}
Scenario                       & Model        & \multicolumn{1}{l}{Threshold} & \multicolumn{1}{l}{Prec} & \multicolumn{1}{l}{DPR} & \multicolumn{1}{l}{$F_{0.5}$} \\ \hline
\multirow{3}{*}{Balanced}  & MLM Bernice  & 50                            & 0.14                     & 6.58                    & 0.17                     \\
                           & MLM BERTweet & 50                            & \textbf{0.23}            & \textbf{9.47}           & \textbf{0.28}            \\
                           & EPD SpanBERT & 50                            & 0.00                     & 0.00                    & 0.00                     \\ \hline
\multirow{3}{*}{Synthetic} & MLM Bernice  & 400                           & 0.50                     & 0.84                    & 0.54                     \\
                           & MLM BERTweet & 50                            & \textbf{2.00}            & 0.42                    & 1.14                     \\
                           & EPD SpanBERT & 800                           & 1.85                     & \textbf{3.36}           & \textbf{2.03}            \\ \hline
\multirow{3}{*}{Realistic} & MLM Bernice  & 6400                          & 0.05                     & \textbf{2.21}           & 0.06                     \\
                           & MLM BERTweet & 400                           & 0.50                     & 1.47                    & 0.58                     \\
                           & EPD SpanBERT & 200                           & \textbf{0.58}            & 0.74                    & \textbf{0.61}            \\ \hline
\end{tabular}%
}
\caption{Results for best MLM and EPD models by $F_{0.5}$. Prec is Precision. DPR is Data Potential Recall. Best scores for across each dataset are \textbf{bolded}}\label{tab:mlmepd}
\end{table}

Table \ref{tab:mlmepd} presents the top-performing MLM and EPD models for each scenario. Overall, MLM and EPD models perform extremely poorly across all metrics. They obtain less than $1$ point in F-score, very poor precision around $1$-$2$ points, and a modest DPR around $1$-$9$ points. For the MLM models, we attribute this lackluster performance to limitation of the tokenizer of the models as MLM models are restricted to predicting single tokens learned from their training corpora. As for the EPD model, we believe that the filtering step using a unigram Word2Vec model leaves out complex phrases. 

Most of the models only predict 50 words mimicking failures in Word2Vec and Phrase2Vec. Additional figures detailing the choice of threshold are provided in Appendix \ref{sec:MLM-EPD-app}. Overall, MLM and EPD models lag behind the simpler embedding models. 

\subsection{EarShot}
\subsubsection{PREDICT}

\begin{table*}[ht]
\resizebox{\linewidth}{!}{%
\begin{tabular}{l|llrrrrr}
Scenario                   & Filtering Model & Keyword Extraction Model & \multicolumn{1}{l}{Threshold} & \multicolumn{1}{l}{Max $n$-grams} & \multicolumn{1}{l}{Prec} & \multicolumn{1}{l}{DPR} & \multicolumn{1}{l}{$F_{0.5}$} \\ \hline
\multirow{2}{*}{Balanced}  & RoBERTA R4      & KeyBERT                  & 50                            & 1                           & \textbf{14.81}           & 1.65                    & \textbf{5.70}           \\
                           & LLaMa 13B       & KeyBERT                  & 100                           & 1                           & 11.43                    & 1.65                    & 5.22                    \\ \hline
\multirow{2}{*}{Synthetic} & RoBERTA R4      & KeyBERT                  & 800                           & 1                           & 14.86                    & \textbf{13.45}          & \textbf{14.55}          \\
                           & LLaMA 8B        & KeyBERT                  & 400                           & 1                           & \textbf{19.13}           & 7.14                    & 14.32                   \\ \hline
\multirow{2}{*}{Realistic} & HateXplain BERT & YAKE                     & 3200                          & 3                           & \textbf{10.00}           & 1.47                    & \textbf{4.63}           \\
                           & LLaMa 13B       & RAKE                     & 200                           & 1                           & 9.52                     & 1.47                    & 4.55                    \\ \hline
\end{tabular}%
}
\caption{Best Results for FETCH LLM and BERT PREDICT based on $F_{0.5}$. Prec is Precision. DPR is Data Potential Recall. Best scores for each dataset are \textbf{bolded}}\label{tab:fetch_predict}
\end{table*}

In Table \ref{tab:fetch_predict}, we choose the best LLM and BERT filtering model for each dataset based on the $F_{0.5}$ of the system. EarShot-PREDICT obtains higher precision than other systems with around $9$-$20$ points. However, its DPR suffers obtaining around $1$-$13$ points. While it beats MLM and EPD in terms of DPR, Word2Vec and Phrase2Vec beat PREDICT on Balanced and Synthetic and tie it on Realistic stemming from keyword extraction models, which aim to identify important words rather than specifically targeting dog whistles themselves. 

To achieve this high precision, the system makes the conservative choice of predicting on average less than 1000 words and only chooses $1-$grams. While its prediction abilities are still restricted, this prediction threshold improves over the other systems as it maintains high precision.

Looking towards the choices of keyword extractors, KeyBERT does the best on smaller datasets like Balanced and Synthetic and YAKE and RAKE do better on larger datasets like Realistic.  For the LLM model choices, LLaMa 13B does the best across Balanced and Realistic while LLaMa 8B does better on Synthetic. As for the BERT model choices, RoBERTa R4 does best across Balanced and Synthetic while HateXplain BERT does well on Realistic. Figures detailing choices of keyword extraction models, filtering models, and thresholds can be found in Appendix \ref{sec:Earshot-Predict-app}.

BERT-based models outperform LLMs by 0.1–0.5 points across all three datasets. Despite their extra parameters, larger LLMs do not surpass task-specific models in this task. 

\subsubsection{DIRECT}

\begin{table}[!ht]
\resizebox{\columnwidth}{!}{%
\begin{tabular}{l|lrrr}
Scenario                    & Model      & \multicolumn{1}{l}{Prec} & \multicolumn{1}{l}{DPR} & \multicolumn{1}{l}{$F_{0.5}$} \\ \hline
\multirow{3}{*}{Balanced}  & LLaMa 13B  & \textbf{2.97}            & 13.58                   & \textbf{3.52}            \\
                           & LLaMa 8B   & 2.35                     & \textbf{21.40}          & 2.86                     \\
                           & Mistral 7B & 2.87                     & 14.81                   & 3.42                     \\ \hline
\multirow{3}{*}{Synthetic} & LLaMa 13B  & \textbf{20.31}           & 56.30                   & \textbf{23.29}           \\
                           & LLaMa 8B   & 14.63                    & 46.64                   & 16.95                    \\
                           & Mistral 7B & 18.26                    & \textbf{58.82}          & 21.18                    \\ \hline
\multirow{3}{*}{Realistic} & LLaMa 13B  & 0.00                     & 0.00                    & 0.00                     \\
                           & LLaMa 8B   & \textbf{0.94}            & \textbf{60.29}          & \textbf{1.17}            \\
                           & Mistral 7B & 0.69                     & 46.32                   & 0.86                     \\ \hline
\end{tabular}%
}\caption{Results for DIRECT Pipeline. Prec is Precision. DPR is Data Potential Recall. Best scores for each dataset are \textbf{bolded}}\label{tab:DIRECT}
\end{table}

Table \ref{tab:DIRECT} summarizes the performance of the EarShot DIRECT system. Direct prompting of an LLM achieves high DPR but moderate precision, due to the model overestimating the number of dog whistles, inflating DPR at the cost of precision.

Diving into the models, LLaMa 13B does the best across Balanced and Synthetic while it fails on Realistic where LLaMa 8B does the best as LLaMa 13B overestimates the number of dog whistles.

DIRECT outperforms all except PREDICT, which surpasses DIRECT on Balanced and Realistic datasets, while DIRECT excels on Synthetic in terms of $F_{0.5}$. DIRECT, though more powerful, fails in realistic settings. However, this result has caveats: DIRECT achieves significantly better DPR, identifies many dog whistles, is less conservative than PREDICT, and is not constrained by an $n$-gram threshold due to its generative nature. 

\section{Discussion}
We evaluate our best EarShot systems across datasets, highlighting their strengths and limitations. We differences between models and analyze challenging dog whistles to steer potential improvements beyond EarShot. We also provide analysis of novel dog whistles found by our system in Appendix \ref{sec:novel-dog-whistles} to analyze putting our method to the test in a simulated moderation scenario.
\subsection{Difference in Dog Whistles}
\subsubsection{Synthetic}
Comparing our best model to the Unigram Word2Vec model, the Word2Vec model finds very simplistic dog whistles like ``pepe'', ``autogynophilia'', ``13/50'', ``npc''. These predictions are limited to internet slang. However, our method is able to capture more complex phraseology like ``middle class'', ``climate alarmists''``entitlement program'' and more often used by politicians. Comparing to the bigram Phrase2Vec, the Phrase2Vec model still chooses simple internet slang like ``amerimutt'' and ``cuckolds''. This shows that in scenarios with many dog whistles, more complex models are able to fetch more dog whistles than Word2Vec. 
\subsubsection{Realistic}
In the Realistic setting, our method achieves high precision, distinguishing dog whistles from noise. It identifies only explicit terms like ``illegal aliens,'' while Word2Vec found some (e.g., ``pedophilia,'' ``groomer'') but struggles with false positives. Moreover, Word2Vec primarily retrieves variants of seed terms (``grooming'', ``pedes''), whereas our method uncovers deeper links (e.g., ``mass migration'' and ``illegal aliens,''), highlighting stronger models in real-world scenarios, where Word2Vec fails.
\subsubsection{Balanced}
Every system has limitations. In the Balanced scenario, Word2Vec identifies subtler and novel dog whistles (e.g., ``gibsmedats'', ``we wuz kangs'', ``Baltimore''  ``chosenites'', ``africanization'', ``vikangs'', ``regressives''), while our method focuses on explicit slurs (``kike'', ``nigger'', ``faggotry.''). This likely results from our conservative threshold, which prioritizes performance, and a BERT filtering system that surfaces only the most overt dog whistles.

\subsection{Difficult Dog Whistles}
First, emoji-based dog whistles were difficult to find as no model detected symbols like the "OK" white power symbol, the milk emoji, or the SS logo as emoji due to the challenges in tokenizing emojis, especially in models with smaller tokenizers.

Second, context dependency and deliberate ambiguity posed significant challenges. Phrases such as ``Barack Hussein Obama,'' ``Federal Reserve,'' ``New York Elite,'' ``Willie Horton,'' and "abolish birthright citizenship" were missed due to their varied contextual usage.

Third, recency presented an issue. Newer dog whistles, like ``jogger'' (coined in 2020 following the shooting of Ahmaud Arbery), ``save women's sports;'' (an anti-trans rallying cry), and ``vegan cat'' (an anti-trans dog whistle), were harder to detect due to evolving and shifting lexicons. Our best models usually defaulted to older and more known dog whistles. In the Realistic scenario, terms like ``illegal immigrants'' and ``illegal aliens,'' were detected. The Balanced scenario identified dog whistles like ``ZOG,'' ``Baltimore,'' ``Sharia,'' and ``thug.'' In the Synthetic scenario, older dog whistles such as ``George Soros,'' ``Google,'' ``Rothschilds," and ``White Lives Matter'' were predicted. However, in the Synthetic scenario, our model did identify some newer dog whistles  like ``genderist'' and ``You Will Never Be a Woman,'', demonstrating promise in detecting emerging terms. 

\subsection{Future Directions}
Our analysis highlights critical areas for enhancing dog whistle discovery. First, complex, nuanced, and recent phrases frequently escape discovery. Second, the dependence on hate speech classifiers tends to prioritize explicit slurs while overlooking subtler forms of harmful language. Third, the inherent precision-recall tradeoff presents substantial challenges, as evidenced by our system’s high precision and Word2Vec’s superior recall. 

Additional improvements could include hybridizing Word2Vec and LLMs, post-processing noisy predictions to enhance precision, ensembling multiple LLMs, scaling model size, adding chain-of-thought reasoning \cite{wei2023chainofthoughtpromptingelicitsreasoning}, or integrating syntactic or semantic information.

\section{Conclusion}
This work lays the groundwork for detecting novel dog whistles with NLP. We introduce a new evaluation task, rigorously test state-of-the-art systems across multiple datasets, and propose a superior approach. While Word2Vec and BERT-based systems under perform, our method shows promise in discovering new dog whistles but needs further development for scalability and comprehensive detection. We hope FETCH! inspires future research in dog whistle discovery and advances NLP's capabilities.

\section{Ethical Considerations}
We identify three key ethical considerations for our work. First, dog whistles are highly culturally specific. Since our study focuses on American cultural contexts, our system may misclassify words or phrases that are not considered dog whistles in other cultures, limiting its applicability and risking false positives when applied cross-culturally. Second, our system is designed to detect hidden or coded language. This ability could have the unintended consequence of exposing language that serves to protect activism or marginalized groups. Third, there is a risk that our system may exhibit bias, disproportionately classifying language from minority communities as dog whistles. Similar systems for hate speech detection have historically encountered fairness challenges \cite{sap-etal-2019-risk}. 

While we acknowledge these risks, we argue that deploying the system in real-world settings, with appropriate safeguards, can effectively mitigate potential harms. Specifically, we recommend incorporating human oversight during deployment to review model outputs, ensuring accountability, and minimizing the risks of false positives and biases. Furthermore, given the critical importance of addressing hate speech, we contend that the benefits of deploying a carefully managed system like ours outweigh the potential harms when implemented responsibly.

\section{Limitations}
We identify a couple key limitations of our paper. First, we work with LLMs that are expensive to run requiring fairly large GPUs to run experiments with. We try to limit ourselves to smaller but still high performance open source models in order to reduce the overhead of our system. 

Second, while a majority of dog whistles are fairly unique and can be found using regex, the lack of a dedicated human labeled corpora hinders the accuracy of the benchmark as false positives could occur. We aimed to limit this problem by including the Synthetic scenario, where the dog whistles were found through different means outside of regex. 

Third, data contamination may pose a concern, as LLMs trained on internet-scale datasets could have encountered the dog whistle dataset during pretraining, potentially inflating their apparent performance. More broadly, the models might have encountered these dog whistles in unlabeled contexts. Consequently, we cannot ensure that the models have not been exposed to these examples prior to evaluation.

Finally, our method has only been tried on English only datasets. However, dog whistles have not been studied in multilingual contexts as much as in English and there are less resources to study them.

\bibliography{custom,anthology}
\bibliographystyle{acl_natbib}

\onecolumn
\appendix

\section{Dog Whistles Found in Each Scenario}\label{appx:scenario-dogwhistles}

\begin{longtable}{|l|p{14cm}|}
\hline
\textbf{Scenario} & \textbf{List of Dog Whistles} \\ \hline
         Balanced & \#genderwoowoo,\#milk,\#womenwontwheesht,109,1290,1488,23/16,absent fathers,affirmative action,aiden,alarmism,alarmist,all lives matter,alt-right,america first,amerimutt,amish,anointed,attacks on or imagery of jewish financial leaders,attacks on or imagery of jewish political leaders,autoandrophile,autogynephile,baby daddies,baby mama,back the blue,bankers,banksters,barack hussein obama,bear emoji,beta,big government,big pharma,bing,biological man,biological realism,black-on-black crime,blue lives matter,blueish,boogaloo,bop,broken family,burning coal,butterfly,cabal,checkered flag emoji,china virus,clown world,clownfish,coin,colorblind,controlled media,critical race theory,cuck,cultural enrichment,cut taxes,deadbeat dad,deep state,dinosaur emojis,durden,election integrity,entitlement programs,entitlement spending,every single time,fafo,family values,federal reserve,food stamps,fren,frog emoji,gangbanger,gender critical,gender ideology,george soros,ghetto,gibsmedat,globalism,globalist,goody,google,government handout,groomers,hard-working americans,he wuz a good boy,hollywood elite,identify as,illegal aliens,illegal immigrant,illuminati,inner city,international bankers,islamic extremism,islamic extremists,islamic terrorism,islamic terrorists,islamists,israel lobby,"its okay to be white",jogger,judas,jwoke,kek,khazars,law and order,lesbophobia,"lets go brandon",lgb rights,lifelong bachelor,lizard emoji,maga,male violence,milk emoji,multiculturalism,neoliberal,new world order,nibba,npc,ok sign emoji,our guy,overpopulation,oy vey,pepe the frog,personal responsibility,pine tree emoji,political correctness,power level,prefix "super",public school,pulling strings,puppet masters,quotas,race realism,radical islam,reagan,red square emoji,references to cities with large racial minority populations being overrun by crime, drugs, rodents,religious freedom,right to work,rothschilds,safeguarding,school choice,secure the border,send me,sharia law,shlomo,silent majority,single parent,skittle,skype,snowflake,sonnenrad,soy boy,special interests,spiderweb emoji,spqr,steroids,strapping young buck,take back,thug,tough on crime,trans identified female,trans identified male,tras,triple parentheses,two lightning bolt emojis,vinland,voter fraud,war on christmas,war on drugs,welfare,white lives matter,windmill,working class,xx,yahoo,ykw,zionist,zionist occupation government\\ \hline
         Synthetic & social justice warrior, blueish, LGB rights, working class,Islamic terrorists, centipede, cuck, TRAs, globalist,Trans Identified Male, Reagan, Islamic extremists,Pepe the Frog, single parent, illegal aliens,public school, MAGA, voter fraud, soy boy, XX,gender critical, political correctness, coincidence,cut taxes, grooming, neoliberal, big pharma, Clown World,America First, government handout, New World Order,multiculturalism, references to cities with large racial minority populations being overrun by crime, drugs, rodents,beta, Islamic extremism, quotas, family values,inner city, cabal, religious freedom, cultural Marxists,steroids, NPC, deadbeat dad, 13\%, fellow white people,power level, Zionist, dual citizen, 1488, Skittle,Google, race realism, Deus Vult, puppet masters, Bing,critical race theory, secure the border,Trans Identified Female, OK sign emoji, octopus,personal responsibility, actual woman, entitlement programs,baby mama, cultural enrichment, Judeo-Christian,welfare queen, gender ideology, Rothschilds, war on terror,job creators, school choice, "states rights", anointed,majority-minorit y, shekels, Barack Hussein Obama,Remove Kebab, shoah, anchor babies, ghetto,forced diversity, judicial activism, Sharia law,Federal Reserve, middle class, Hollywood elite, welfare,deep state, George Soros, illegal immigrant,affirmative action, snowflake, thug, food stamps,radical Islam, YKW, take back, law and order,the goyim know, cultural Marxism, right to work, globalism,mass immigration, bankers, genderfree, election integrity,goy, gibsmedat, Islamic terrorism, bugman,black-on-black crime, strapping young buck, biological man,frog emoji, alt-right, Kek, womyn, silent majority,Islamists, war on drugs, autogynephile, biological realism,lifelong bachelor, groomers, based, identify as,big government, entitlement spending, our guy, Yahoo,fren, "its okay to be white", Skype,institution of marriage, every single time, absent fathers,biological woman, dindu nuffin, Blue Lives Matter, Amish,special interests, triple parentheses, 41\%, Pajeet,agender, hard-working Americans, forced busing,out of wedlock, broken family, property rights,safeguarding, balance the budget, international bankers,alarmist, right to know, Illuminati, male violence,overpopulation, Jwoke, All Lives Matter, autoandrophile,pulling strings, "Lets Go Brandon", baby daddies, oy vey,cosmopolitan, freedom of association, Judas,war on Christmas, hygienic, colorblind, fatherless,identitarianism, China Virus, gender socialization,controlled media, White Lives Matter, bop, early life,poisoning the well, sex is real, gangbanger,black and orange square emojis, health freedom,Zionist Occupation Government, we wuz kangz, windmill,global elite, Kalergi Plan, dindu, preemptive strike,mixed sex, economic anxiety, car salesman, genderist,brave and stunning, jogger, prefix "super", thin blue line,Sonnenrad, banksters, troon, Trilateral Commission,vaccine safety, implicit bias, central bankers,healthy tissue, erasing women, he wuz a good boy,sex-based rights, lesbian erasure, sex not gender,lesbophobia, intact, RWDS, alarmism,rootless cosmopolitan, magapede, surgical wound,burning coal, Three Percenters, gender abolitionist,three strikes laws, superpredators, SPQR,attacks on or imagery of Jewish political leaders, Dred Scott,welfare cheats, coal burner, uterus-haver, gender fandom,string puller, Boogaloo, pine tree emoji, lizard emoji,New York elite, \#GenderWooWoo, New York intellectuals,protecting women and girls, merit-based immigration policy,YWNBAW, dinosaur emojis, milk emoji, "save womens sports",biological realist, cleaning up our streets, 109, COIN,Amerimutt, tough on crime, Goody, autogynephilia, Durden,coastal elite, vegan cat, FAFO, physical removal,Khazars, Aiden, pro-family, Vinland, dual loyalty,SWPL, womanface, Willie Horton, bear emoji,attacks on or imagery of Jewish financial leaders, pilpul,A Leppo, cosmopolitan elite, Holocauster, Carolus Rex,loxism, free helicopter rides, Back the Blue,adult human female, clownfish, gender abolition,abolish birthright citizenship, bix nood, war on crime,RaHoWa, cherry emoji, two lightning bolt emojis,checkered flag emoji, autoandrophilia, AFAB trans woman,butterfly, Israel Lobby, nibba, male-friendly content,open borders for Israel, Shlomo, tiny minority of men,1290, New York values \\ \hline
         Realistic & \#genderwoowoo,\#milk,\#womenwontwheesht,109,1290,1488,23/16,absent fathers,affirmative action,aiden,alarmism,alarmist,all lives matter,alt-right,america first,amerimutt,amish,anointed,attacks on or imagery of jewish financial leaders,attacks on or imagery of jewish political leaders,autoandrophile,autogynephile,baby daddies,baby mama,back the blue,bankers,banksters,barack hussein obama,bear emoji,beta,big government,big pharma,bing,biological man,biological realism,black-on-black crime,blue lives matter,blueish,boogaloo,bop,broken family,burning coal,butterfly,cabal,checkered flag emoji,china virus,clown world,clownfish,coin,colorblind,controlled media,critical race theory,cuck,cultural enrichment,cut taxes,deadbeat dad,deep state,dinosaur emojis,durden,election integrity,entitlement programs,entitlement spending,every single time,fafo,family values,federal reserve,food stamps,fren,frog emoji,gangbanger,gender critical,gender ideology,george soros,ghetto,gibsmedat,globalism,globalist,goody,google,government handout,groomers,hard-working americans,he wuz a good boy,hollywood elite,identify as,illegal aliens,illegal immigrant,illuminati,inner city,international bankers,islamic extremism,islamic extremists,islamic terrorism,islamic terrorists,islamists,israel lobby,"its okay to be white",jogger,judas,jwoke,kek,khazars,law and order,lesbophobia,"lets go brandon",lgb rights,lifelong bachelor,lizard emoji,maga,male violence,milk emoji,multiculturalism,neoliberal,new world order,nibba,npc,ok sign emoji,our guy,overpopulation,oy vey,pepe the frog,personal responsibility,pine tree emoji,political correctness,power level,prefix "super",public school,pulling strings,puppet masters,quotas,race realism,radical islam,reagan,red square emoji,references to cities with large racial minority populations being overrun by crime, drugs, rodents,religious freedom,right to work,rothschilds,safeguarding,school choice,secure the border,send me,sharia law,shlomo,silent majority,single parent,skittle,skype,snowflake,sonnenrad,soy boy,special interests,spiderweb emoji,spqr,steroids,strapping young buck,take back,thug,tough on crime,trans identified female,trans identified male,tras,triple parentheses,two lightning bolt emojis,vinland,voter fraud,war on christmas,war on drugs,welfare,white lives matter,windmill,working class,xx,yahoo,ykw,zionist,zionist occupation government\\ \hline
\end{longtable}

\newpage

\section{Dog Whistle Examples}\label{appx:dog-whistles-examples}
Dogwhistles are linguistically challenging and fairly niche use of words. We provide more examples from each dataset in context with explanations. 
\subsection{Balanced (Gab)}
\paragraph{Soros}
\begin{quote}
    {
    ``\textit{red flag warning for texas!!! gov abbott: "all americans concerned about losing our liberty to the leftist liberal agenda need to know about \textbf{soros} spending big in texas.}'' \\\phantom{abc}--- \textbf{Anonymous Gab User}
    }
\end{quote}

In this quote, an anonymous user on Gab informs the platform of the spending by billionaire George Soros on liberalizing reforms. Soros is a notorious boogeyman in the right-wing spaces due to his Open Society Foundations, a very influential network of non-profits that donate to liberal and progressive causes. In addition, his Jewish heritage feeds into popular antisemitic tropes about the Jews controlling the world. This dog whistle slips through many filters due to the its ability to mask possible antisemitism under the guise of criticizing billionaire donations. 

\paragraph{Echoing}
\begin{quote}
    {
    ``\textit{from what i read it was not primarily christian reasons jews were expelled, but for practical reasons that \textbf{echo} today's reasons.}'' \\\phantom{abc}--- \textbf{Anonymous Gab User}
    }
\end{quote}

In this quote, an anonymous user on Gab informs the platform of their research on the many expulsions of the Jews from countries during history. The dog whistle here is the use of echo. While echo looks extremely benign in its usage, there are more sinister reasons to choose echo over other similar words. Echo is related to the antisemitic use of the triple parenthesis ((())) around Jewish names, topics, or names of organizations thought to be owned by Jews. These triple parentheses represent that the Jewish people's actions "echo throughout history" hence the use of the term echo. 

\paragraph{Gibs Me Dat}
\begin{quote}
    {
    ``\textit{the solution to rotten tomatoes \textbf{gibs me dat} 100\% black panthers \#blaxit lawerewolf \#gabfam \#maga \#news \#auspol \#speakfreely \#ausfam \#australia \#usa}'' \\\phantom{abc}--- \textbf{Anonymous Gab User}
    }
\end{quote}

In this quote, an anonymous user on Gab insults African Americans wanting the film Black Panther to have a 100\% percent approval rating on the film rating site Rotten Tomatoes. The dog whistle used here is Gibs Me Dat. Due to its extremely weird spelling and non standard usage, one might pass over it as a possible misspelling or worse dialectical usage. However, this phrase mocks African American Vernacular English, a dialect of English that African American's often use, for being unintelligent sounding despite it having a well-formed grammar and more complex rules than Standard American English. 
\subsection{Synthetic (Reddit)}
\paragraph{Baltimore}
\begin{quote}
    {
    ``\textit{that's what blm apparently want, no policing of the black community. they're getting it in chicago and \textbf{baltimore} right now, and using their new freedom to run up the body counts to levels not seen since the drug wars of the early 90s.}'' \\\phantom{abc}--- \textbf{Anonymous Reddit User}
    }
\end{quote}

In this quote, an anonymous user on Reddit criticizes the police practices of cities like Chicago and Baltimore and the influence of the Black Lives Matter movement on these decisions. The dog whistle here is Baltimore. This dog whistle is extremely difficult to pick up because it is a regular city name. However, it is a dog whistle because it is used in a similar vein to the use of ``inner cities.'' Politicians reference a place that often has high African American populations like inner cities and Baltimore to disparage the people living in it. On a side note, Baltimore is a wonderful city that you should visit \smiley. 

\paragraph{Coincidences}
\begin{quote}
    {
    ``\textit{funny how there's been tons of rallies before but suddenly there's all sorts of nazi imagery just so convenient for the cameras at this specific rally. probably just a \textbf{coincidence} i'm sure..}'' \\\phantom{abc}--- \textbf{Anonymous Reddit User}
    }
\end{quote}

In this quote, an anonymous user on Reddit believes that the appearance of Nazi imagery at a rally was not a coincidence and someone deliberately placed it there. The dog whistle here is coincidence. Coincidence often references the antisemitic trope that the Jews control the world and that many coincidences are actually Jewish conspiracies. This dog whistle slips under the radar as one could be referring to funny coincidences like being in the same vacation location as your friend at the same time or interesting coincidences like Mark Twain predicting his own death. 

\paragraph{Family Values}
\begin{quote}
    {
    ``\textit{holy shit... nfl is done, \textbf{family values} households are shutting this crap off big-time. thanks for sharing!}'' \\\phantom{abc}--- \textbf{Anonymous Reddit User}
    }
\end{quote}

In this quote, an anonymous Reddit user displays their happiness that people with family values are boycotting the National Football League due to their support of LGBTQ+ rights. The dog whistle here is family values which references the fact that the user does not believe queer life styles are normal. This dog whistle is tricky to find because family values is a catch all term that could have positive connotations but in this specific context, disguises a hatred towards queer relationships. 

\subsection{Realistic (Twitter)}

\paragraph{Islamist}
\begin{quote}
    {
    ``\textit{\textbf{islamist} congresswoman \#ilhanomar is jeopardising american foreign policy so that she can pander to her base.}'' \\\phantom{abc}--- \textbf{Anonymous Twitter User}
    }
\end{quote}

In this quote, an anonymous Twitter user displays their hatred of Congresswoman Ilhan Omar of Minnesota for her foreign policy decisions. They deride her with the term Islamist. Islamism is an academic term which encompasses all political and religious movements that believe Islam should influence politics. However, it hides the more sinister usage of calling Ilhan Omar a terrorist, an extremist, and a traitor to the United States.  

\paragraph{Illegals}
\begin{quote}
    {
    ``\textit{\textbf{illegals} coming here by the thousands every day will drain every resource there is they will over crowd the schools}'' \\\phantom{abc}--- \textbf{Anonymous Twitter User}
    }
\end{quote}

In this quote, an anonymous Twitter user displays their distain for illegal immigrants and their supposed draining of government resources. The dog whistle here is illegals. Illegal immigrants are a diverse group of people ranging from many different places around the world. However, often when people deride illegal immigrants they often talk about Latino immigrants. This meaning often slips by filters because talking about illegal immigrants is not necessarily hate speech, but in some contexts, it is used as a cudgel against Latino people as opposed to other illegal immigrants from other countries.

\section{Prompts used in EarShot}\label{appx:prompts}
\begin{table}[!ht]
    \centering
    \begin{tabular}{|l|p{14cm}|}
    \hline
\textbf{EarShot Method} & \textbf{Prompt} \\ \hline
      LLM-PREDICT   &  "You are a helpful assistant participating in a content moderation task of dog whistle detection.\textbackslash n"
        "A dog whistle is the use of coded or suggestive language in political messaging to garner support from a particular group without provoking opposition. The concept is named after ultrasonic dog whistles, which are audible to dogs but not humans. Dog whistles use language that appears normal to the majority but communicates specific things to intended audiences. They are generally used to convey messages on issues likely to provoke controversy without attracting negative attention. Are dog whistles present in the following sentence?\textbackslash n"
        "\{POST\}\textbackslash n"
        "Please respond ONLY Yes or No. Do not explain your answer.\textbackslash n"\\ \hline
      DIRECT   & "You are a helpful assistant participating in a content moderation task of dog whistle detection.\textbackslash n"
        "A dog whistle is the use of coded or suggestive language in political messaging to garner support from a particular group without provoking opposition. The concept is named after ultrasonic dog whistles, which are audible to dogs but not humans. Dog whistles use language that appears normal to the majority but communicates specific things to intended audiences. They are generally used to convey messages on issues likely to provoke controversy without attracting negative attention. Does the following post contain dog whistles? If they do, please identify the dog whistles.\textbackslash n"
        "\{POST\}\textbackslash n"
        'Please respond with a JSON for a function call with its proper arguments that best answers the given prompt. Respond in the format \{"dogwhistles": list of strings\}.\textbackslash n' \\\hline
    \end{tabular}
    \caption{Prompts used in EarShot}
    \label{tab:prompt_table}
\end{table}

\newpage

\section{Word2Vec/Phrase2Vec Threshold Analysis}\label{sec:word2vec-app}

\begin{figure*}[!ht]
    \centering
    \includegraphics[width=1.05\linewidth]{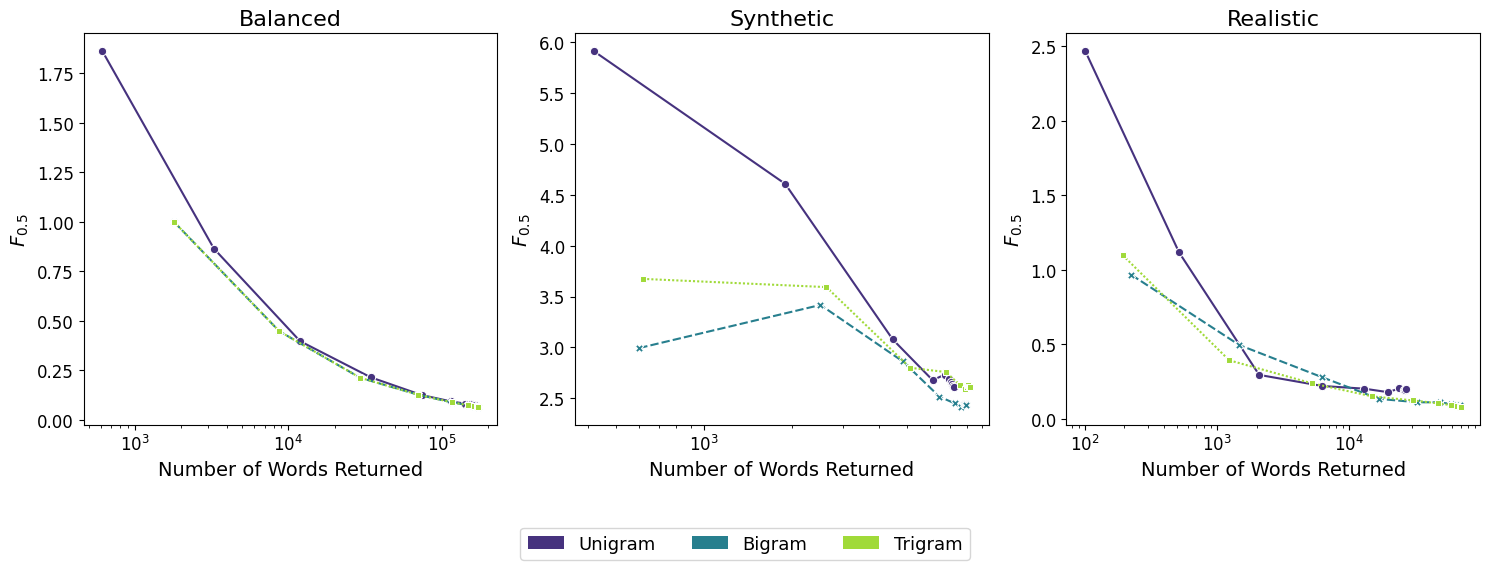}
    \caption{Word2Vec/Phrase2Vec $F_{0.5}$ performance vs the number of words/phrases returned by the model. Plot is on log scale.}
    \label{fig:word2vec}
\end{figure*}

\newpage

\section{MLM and EPD Threshold Analysis}\label{sec:MLM-EPD-app}

\begin{figure*}[!ht]
    \centering
    \includegraphics[width=1.05\linewidth]{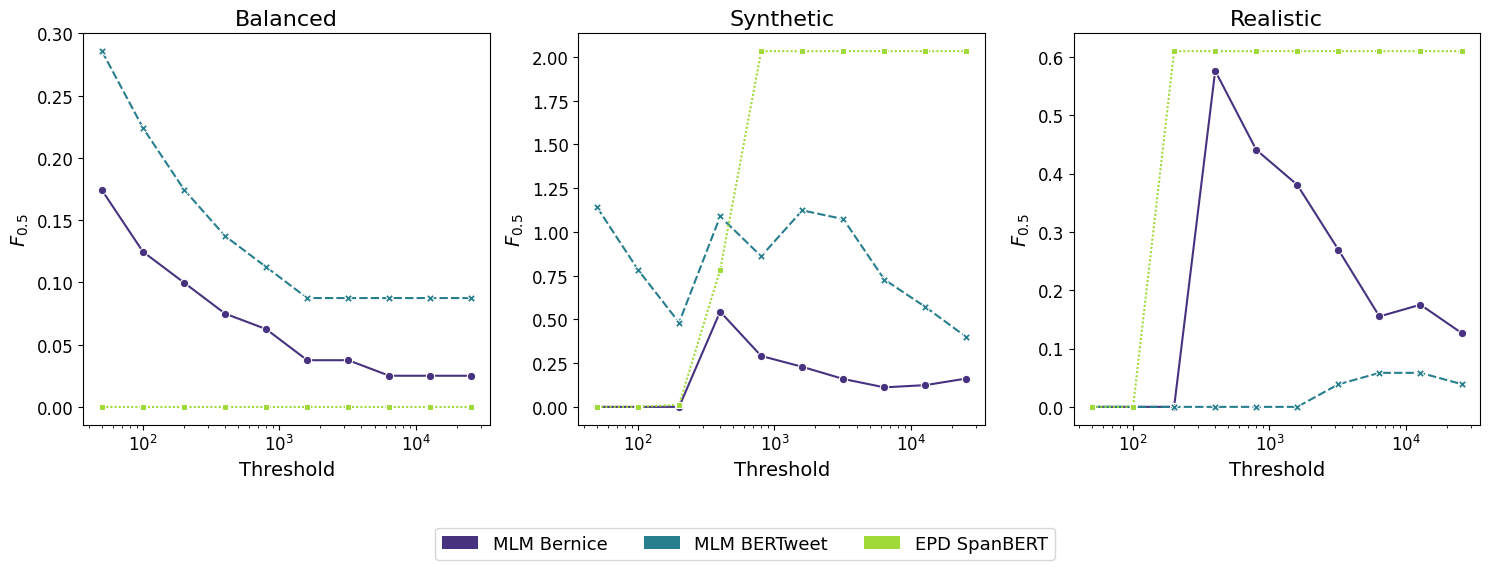}
    \caption{MLM and EPD $F_{0.5}$ performance vs the prediction threshold. Plot is on log scale. }
    \label{fig:mlm-epd}
\end{figure*}

\newpage

\section{EarShot PREDICT Threshold Analysis}\label{sec:Earshot-Predict-app}
\subsection{EarShot BERT PREDICT Threshold Analysis}
\begin{figure*}[!ht]
    \centering
    \includegraphics[width=1.05\linewidth]{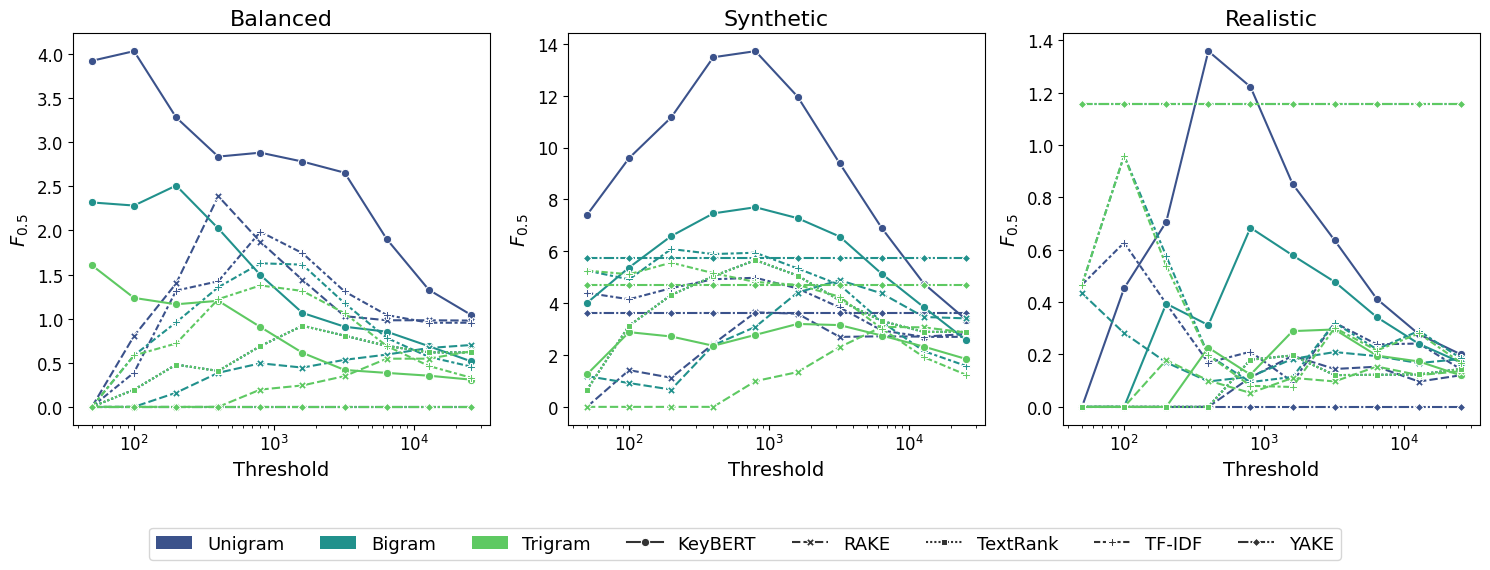}
    \caption{$F_{0.5}$ averaged across BERT based filtering methods displaying average performance across keyword extraction models for EarShot BERT PREDICT vs prediction threshold across all three datasets. Plot is on log scale.}
    \label{fig:fetch-keyword-predict-bert}
\end{figure*}

\begin{figure*}[!ht]
    \centering
    \includegraphics[width=1.05\linewidth]{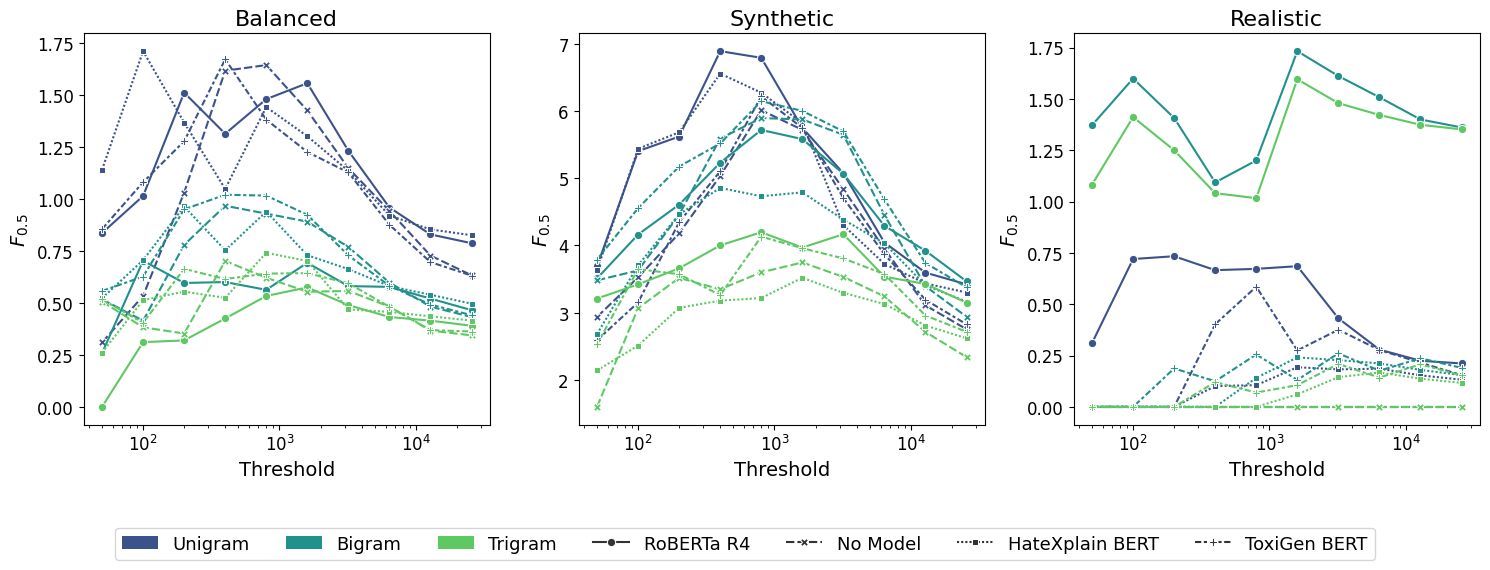}
    \caption{$F_{0.5}$ averaged across keyword extraction models displaying average performance across BERT based filtering models for EarShot BERT PREDICT vs prediction threshold across all three datasets. Plot is on log scale. }
    \label{fig:fetch-model-predict-bert}
\end{figure*}

\newpage

\subsection{EarShot LLM PREDICT Threshold Analysis}
\begin{figure*}[!ht]
    \centering
    \includegraphics[width=1.05\linewidth]{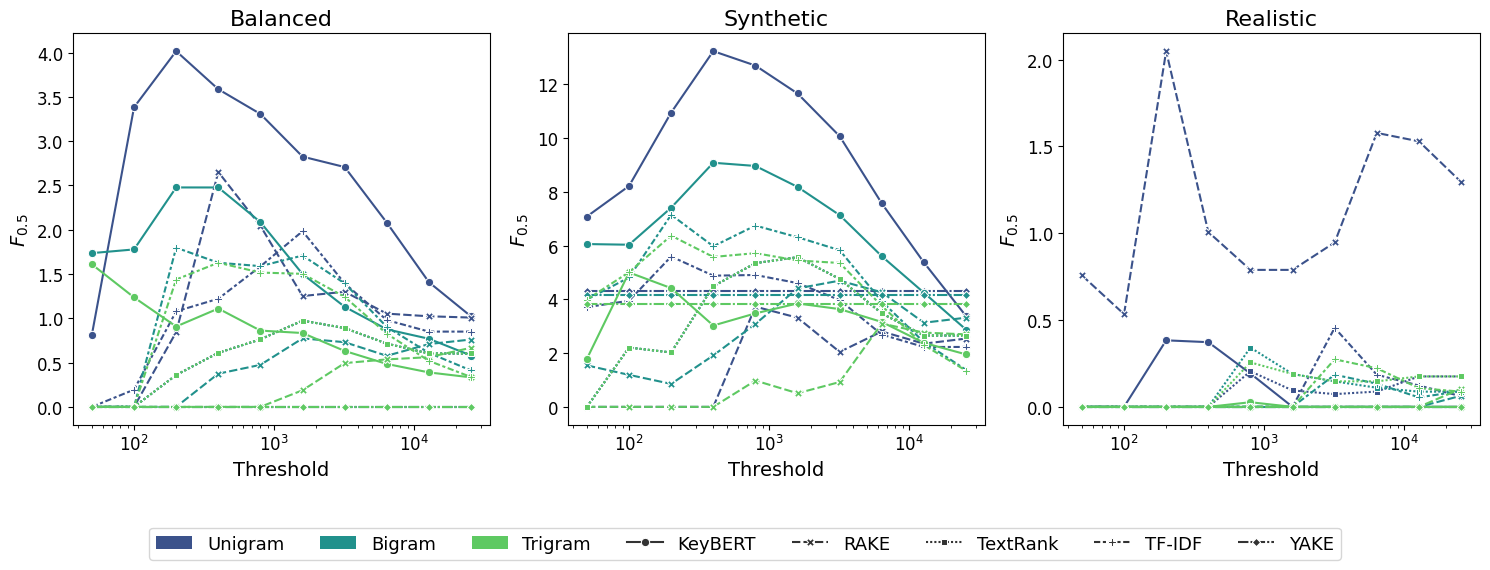}
    \caption{$F_{0.5}$ averaged across LLM based filtering methods displaying average performance across keyword extraction models for EarShot LLM PREDICT vs prediction threshold across all three datasets. Plot is on log scale. }
    \label{fig:fetch-keyword-predict-llm}
\end{figure*}

\begin{figure*}[!ht]
    \centering
    \includegraphics[width=1.05\linewidth]{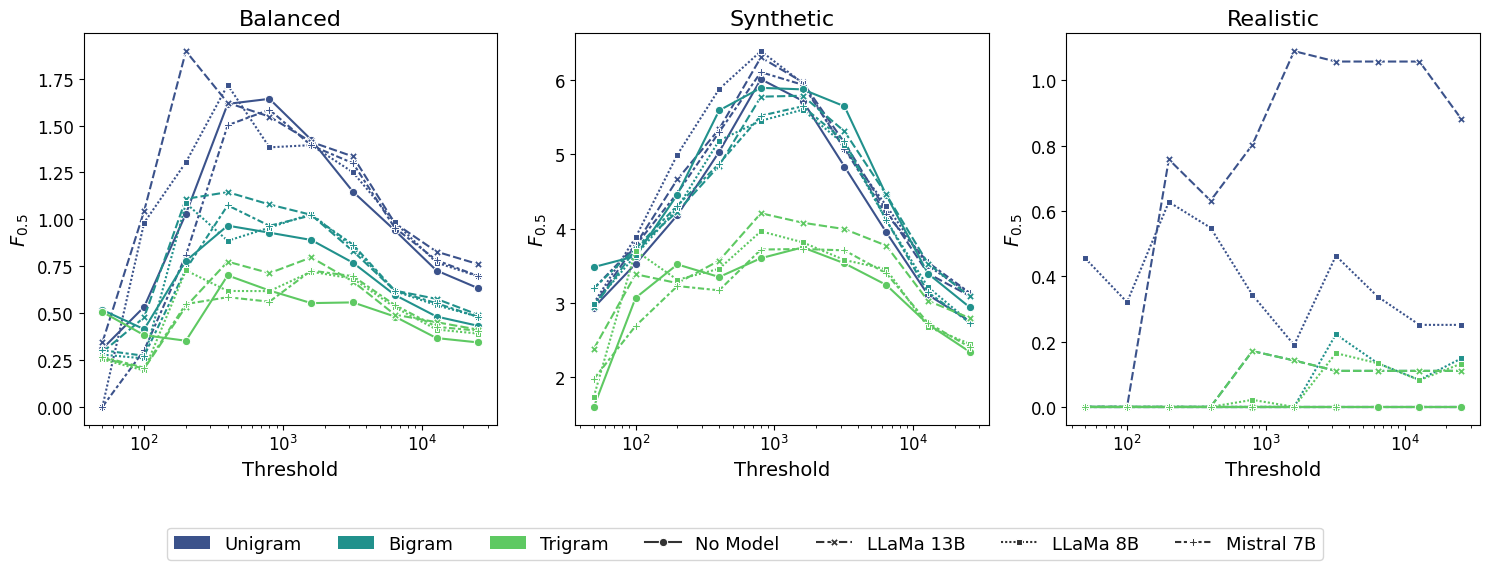}
    \caption{$F_{0.5}$ averaged across keyword extraction models displaying average performance across LLM based filtering models for EarShot LLM PREDICT vs prediction threshold across all three datasets. Plot is on log scale. }
    \label{fig:fetch-model-predict-llm}
\end{figure*}

\section{Novel Dog Whistles}\label{sec:novel-dog-whistles}
We reviewed the top systems' output to identify novel dog whistles in order to measure the effectiveness for dog whistle discovery. We detail 2–3 per dataset with more examples in Appendix \ref{sec:novel-dog-whistles}.
\subsection{Balanced}
The first dog whistle our model identified is "Wakanda," the fictional city from Marvel's Black Panther series. It appears in contexts suggesting African Americans should return to Africa or mocking their enjoyment of the film, as in the post: ``I hate Wakandans,'' where it replaces the N-word. This dog whistle slips under the radar because of its benign initial contexts which could be in a discussion about the films or the comics but when used in this context it could disguise coded racism.  

The second dogwhistle is ``oligarchy.'' Used in the post: ``...the \#DeepState-\#Oligarchy to divide and conquer us. Don't fall for their scam. We're all in this together!'', it ties in with the term ``deep state'' which frequently implies a conspiracy about Jewish global influence. Oligarchy slips under the radar due to its multiple word senses. It could be in the post criticizing the Russian oligarchy, talking about historical oligarchies but also could hide this sense of being antisemitic. 

Another novel dog whistle the model found in the Balanced scenario was brainwashing. Often, this dog whistle is used to describe either a contempt for liberal and progressive thought being supposedly being taught in public schools or Jewish control of the world. For the first meaning, the user talks about hating the supposed liberal indoctrination in schools: ``true. i wouldnt let my kids go to public cesspools. and im so glad. homeschool or private school is the only answer. close the public brainwashing cesspools.'' For the second meaning, explicitly the user talks about the conspiracy of Jewish global control: ``nope! truth isn't brainwashing, it's the antidote to filthy kike lies!'' (kike being a slur for a person of Jewish descent). Brainwashing could slip under the radar as it is used in describing different policies under authoritarian regimes like Nazi Germany or describing life in religious cults. 
\subsection{Synthetic}
One novel dog whistle found from the Synthetic scenario was ``zogbots'', found in the post: ``...turning them into good little zogbots.'' This term refers to the conspiracy theory of a Zionist Occupied Government (ZOG), suggesting Jewish control making people mindless followers.

Another novel dog whistle found in the Synthetic scenario was dindu mcnuffin. It appears in the post ``..impunity while blm is mad that it now shows that dindu mcnuffin did in fact point his stolen gun at the cops/innocent bystanders/puppy.`` This dog whistle is an modification to the previously found version ``dindu nuffin'' which is a contraction of ``did not do nothing'' attempting to sound like African American Vernacular English (AAVE). This phrase mocks AAVE and references the outcry by Black communities after a Black person is a victim of police brutality where they believe that the black person did not do nothing wrong. Dindu mcnuffin could slip under the radar as its modification of dindu nuffin which could throw off classifiers or word lists for detection. In addition, its resemblance to a name also could throw off classifiers.

\subsection{Realistic}
Our best system did not find any new dog whistles due to its conservative threshold.

\newpage

\end{document}